\documentclass[runningheads]{llncs}
\usepackage[T1]{fontenc}
\usepackage{graphicx}
\usepackage{booktabs}
\usepackage{array}
\usepackage{subcaption}
\usepackage{amsmath}
\usepackage{cleveref}
\usepackage{amssymb}
\usepackage{marvosym}
\begin{document}
\title{SwinSF: Image Reconstruction from Spatial-Temporal Spike Streams}
%
%
\author{Liangyan Jiang\inst{1} \orcidID{0009-0001-6934-0527} \and
Chuang Zhu\inst{1} \orcidID{0000-0001-5155-7069} \Letter \and
Yanxu Chen\inst{1} \orcidID{0009-0004-1805-3239}}
\authorrunning{Jiang et al.}
\institute{Beijing University of Posts and Telecommunications, China
\email{\{lander,czhu,cyx666\}@bupt.edu.cn}}
\maketitle              
\begin{abstract}
The spike camera, with its high temporal resolution, low latency, and high dynamic range, addresses high-speed imaging challenges like motion blur. 
   It captures photons at each pixel independently, creating binary spike streams rich in temporal information but challenging for image reconstruction. 
   Current algorithms, both traditional and deep learning-based, still need to be improved in the utilization of the rich temporal detail and the restoration of the details of the reconstructed image.
   To overcome this, we introduce Swin Spikeformer (SwinSF), a novel model for dynamic scene reconstruction from spike streams.
   SwinSF is composed of Spike Feature Extraction, Spatial-Temporal Feature Extraction, and Final Reconstruction Module.
   It combines shifted window self-attention and proposed temporal spike attention, ensuring a comprehensive feature extraction that encapsulates both spatial and temporal dynamics, leading to a more robust and accurate reconstruction of spike streams.
   Furthermore, we build a new synthesized dataset for spike image reconstruction which matches the resolution of the latest spike camera, ensuring its relevance and applicability to the latest developments in spike camera imaging.
   Experimental results demonstrate that the proposed network SwinSF sets a new benchmark, achieving state-of-the-art performance across a series of datasets, including both real-world and synthesized data across various resolutions.
   Our codes and proposed dataset can be found at \url{https://github.com/bupt-ai-cz/SwinSF}.

\end{abstract}

\begin{figure}[tb]
  \centering
  \begin{subfigure}[b]{0.23\textwidth} 
    \centering 
    \includegraphics[height=3cm]{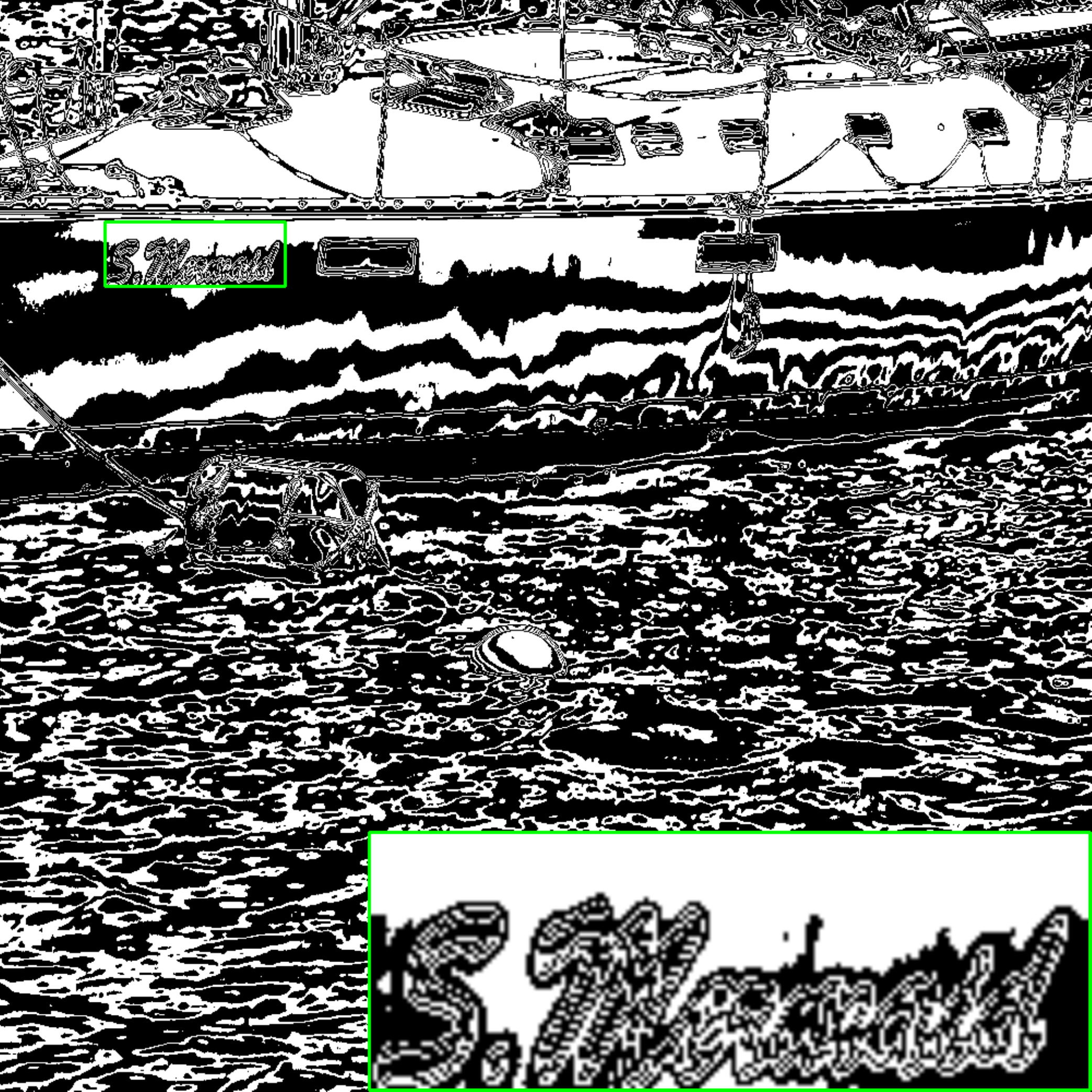}
    \caption{Spike Streams}
    \label{fig:intro-spike}
  \end{subfigure}
  \hfill 
  \begin{subfigure}[b]{0.23\textwidth} 
    \centering 
    \includegraphics[height=3cm]{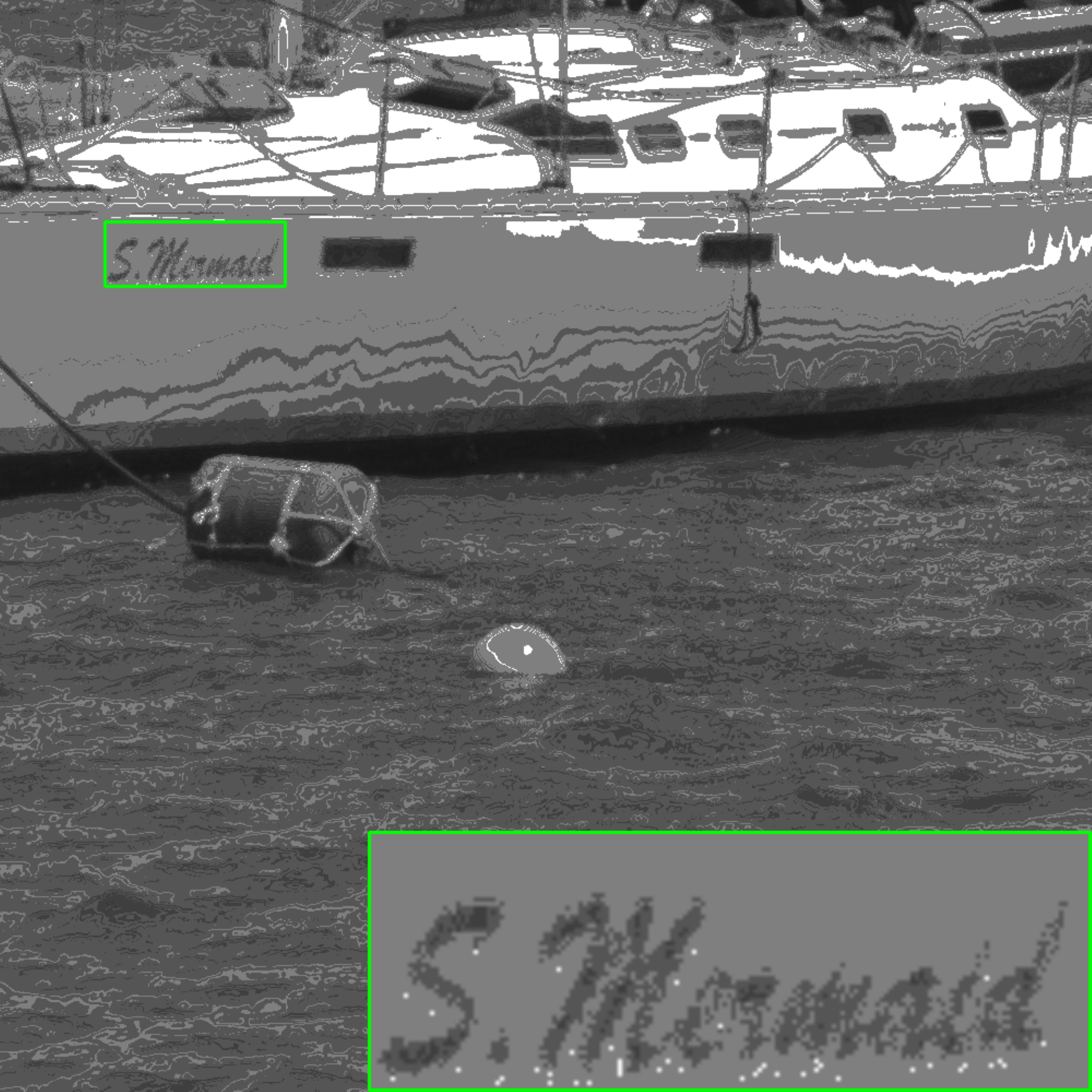}
    \caption{TFI\cite{Alpher02}}
    \label{fig:intro-tfi}
  \end{subfigure}
  \hfill 
  \begin{subfigure}[b]{0.23\textwidth}
    \centering 
    \includegraphics[height=3cm]{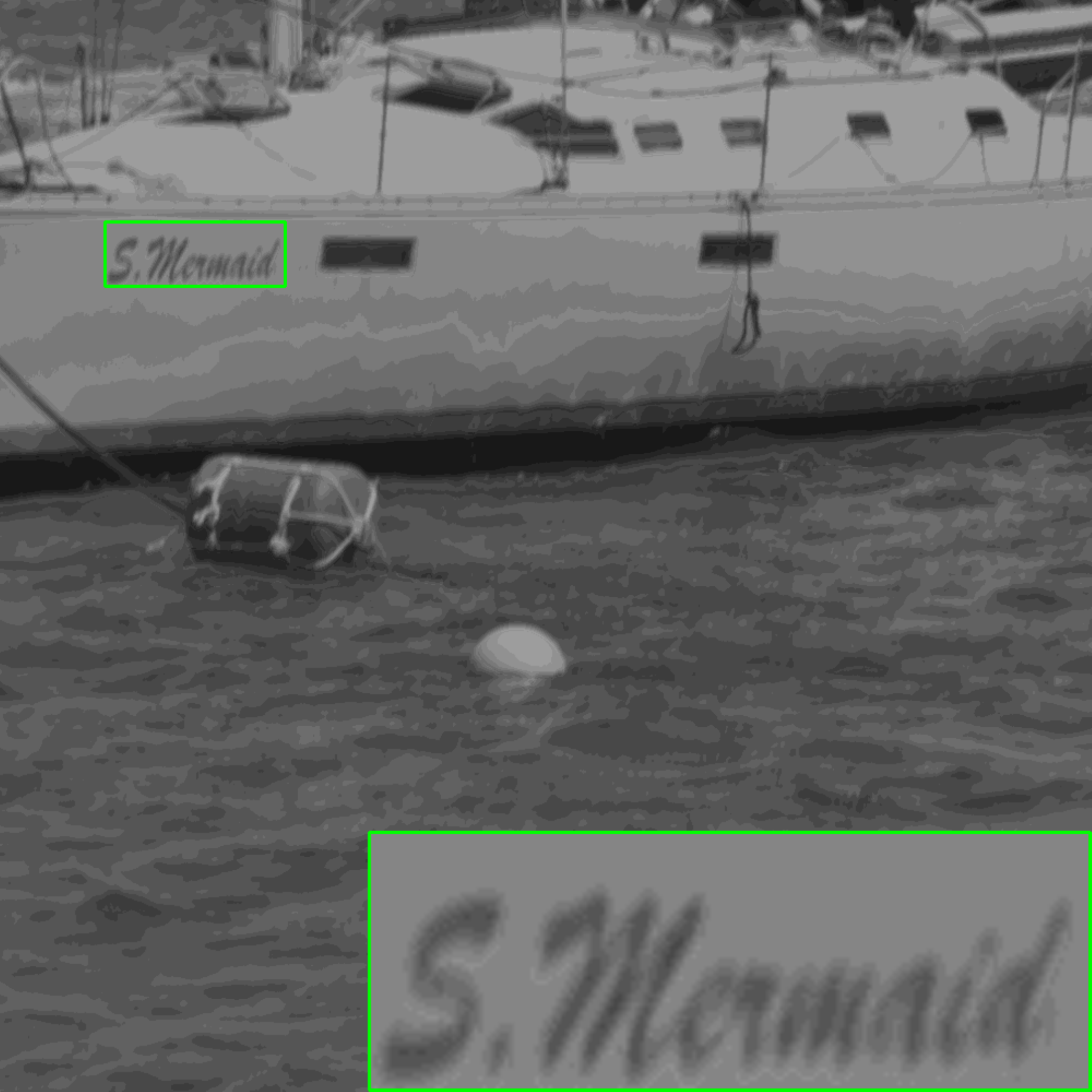}
    \caption{TFP\cite{Alpher02}}
    \label{fig:intro-tfp}
  \end{subfigure}
  \hfill 
  \begin{subfigure}[b]{0.23\textwidth}
    \centering 
    \includegraphics[height=3cm]{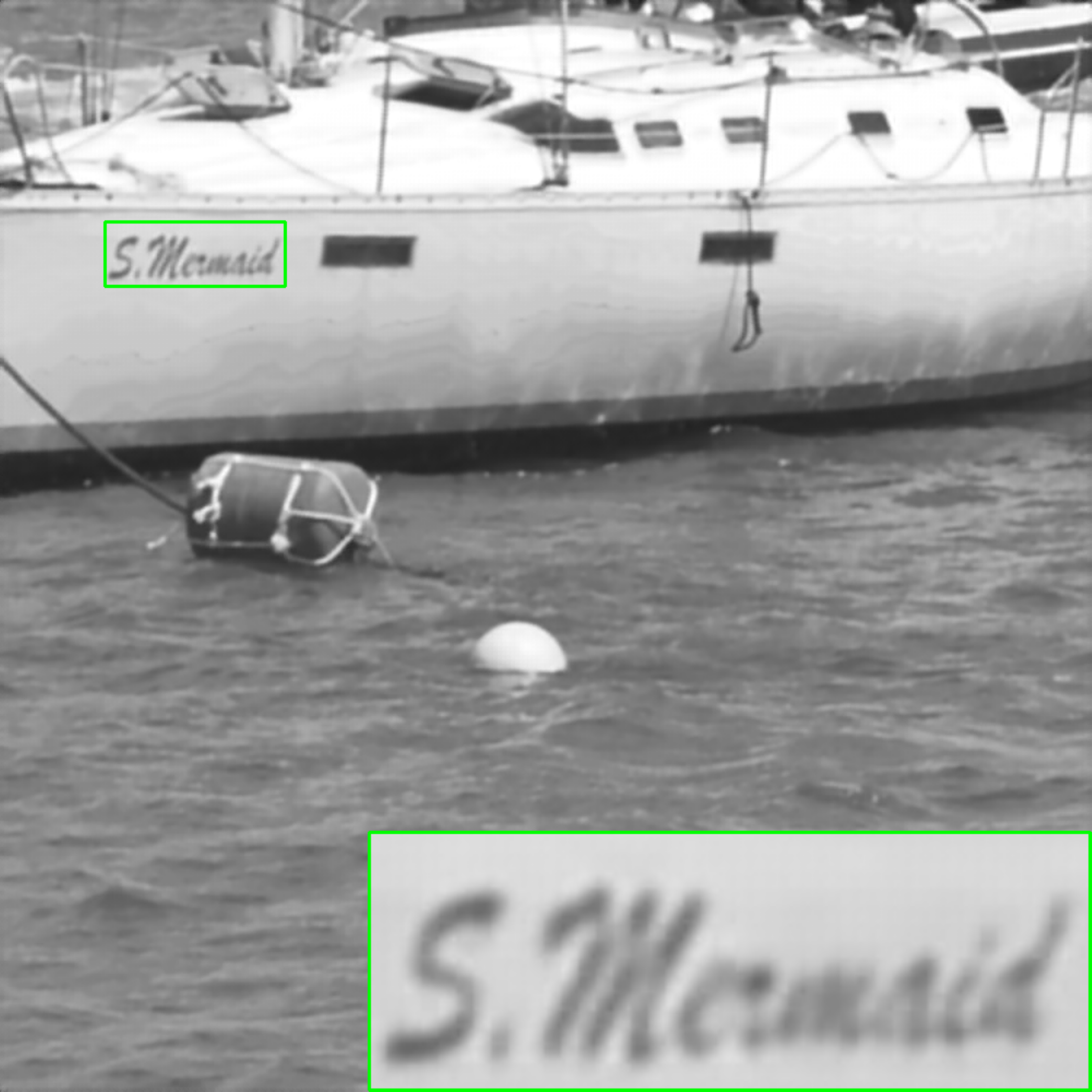}
    \caption{SpikeFormer\cite{Alpher08}}
    \label{fig:intro-spikeformer}
  \end{subfigure}
  \medskip
  \begin{subfigure}[b]{0.23\textwidth} 
    \centering 
    \includegraphics[height=3cm]{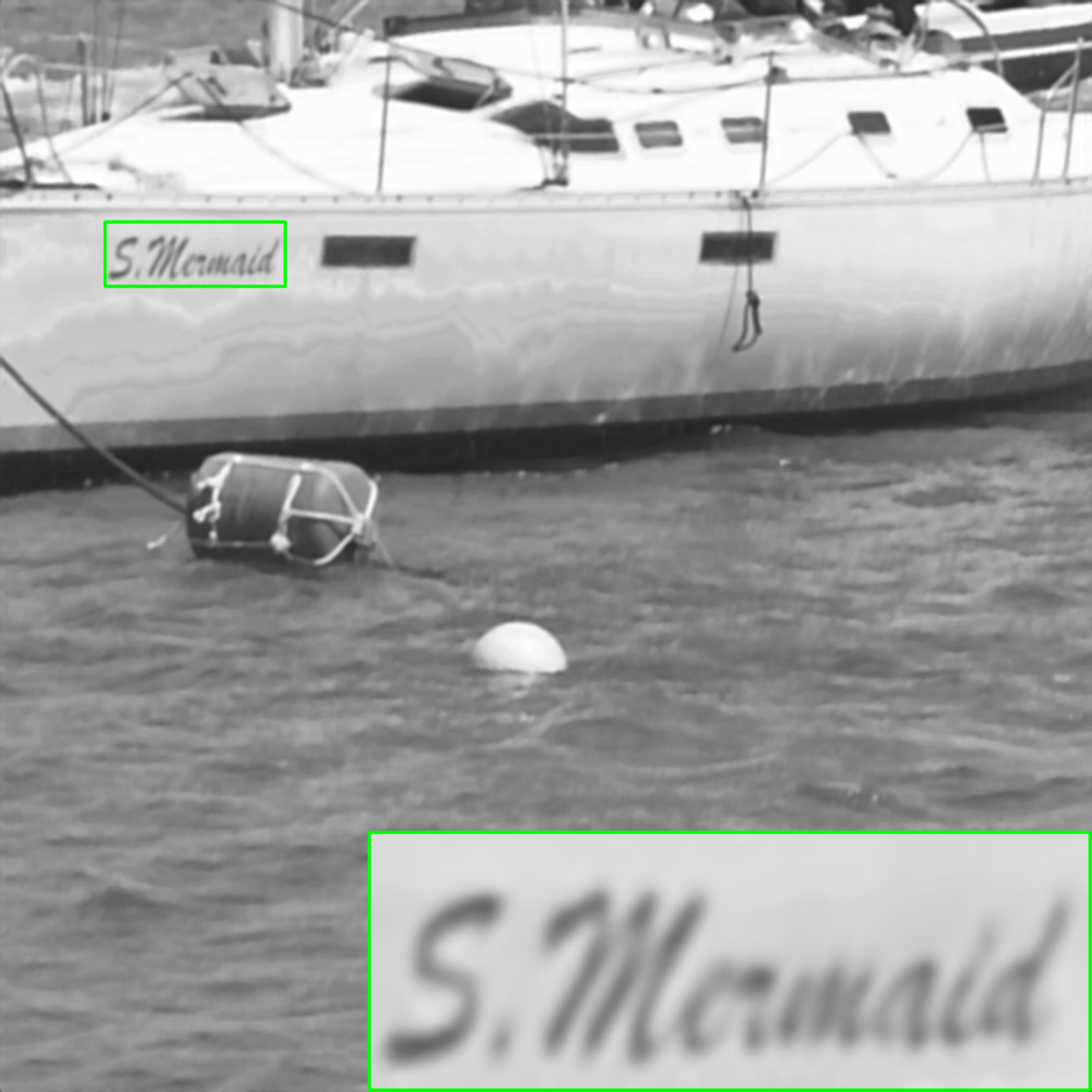}
    \caption{Spk2ImgNet\cite{Alpher07}}
    \label{fig:intro-Spk2ImgNet}
  \end{subfigure}
  \hfill 
  \begin{subfigure}[b]{0.23\textwidth} 
    \centering 
    \includegraphics[height=3cm]{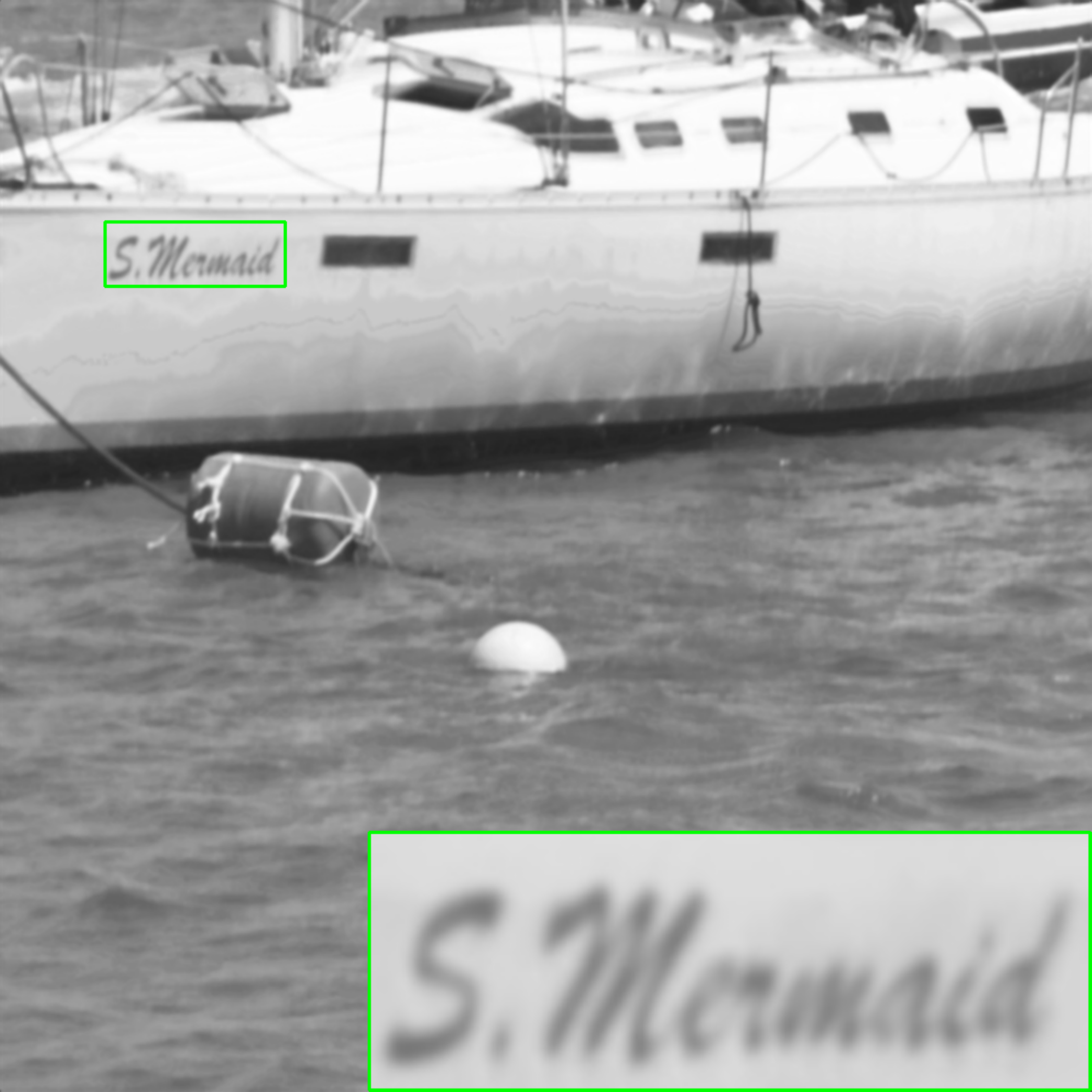}
    \caption{WGSE\cite{Alpher09}}
    \label{fig:intro-WGSE}
  \end{subfigure}
  \hfill 
  \begin{subfigure}[b]{0.23\textwidth}
    \centering 
    \includegraphics[height=3cm]{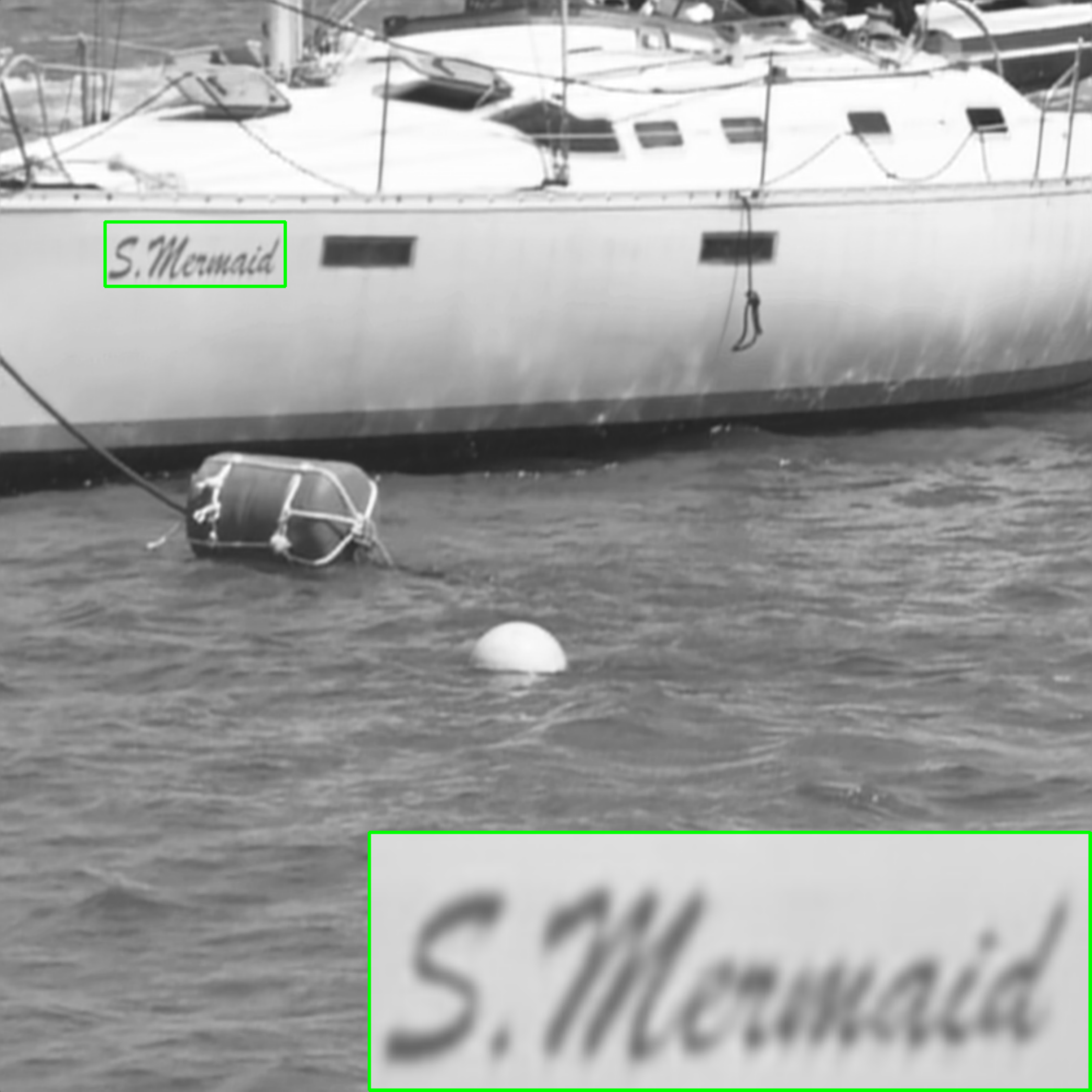}
    \caption{Ours(SwinSF)}
    \label{fig:intro-ours}
  \end{subfigure}
  \hfill 
  \begin{subfigure}[b]{0.23\textwidth}
    \centering 
    \includegraphics[height=3cm]{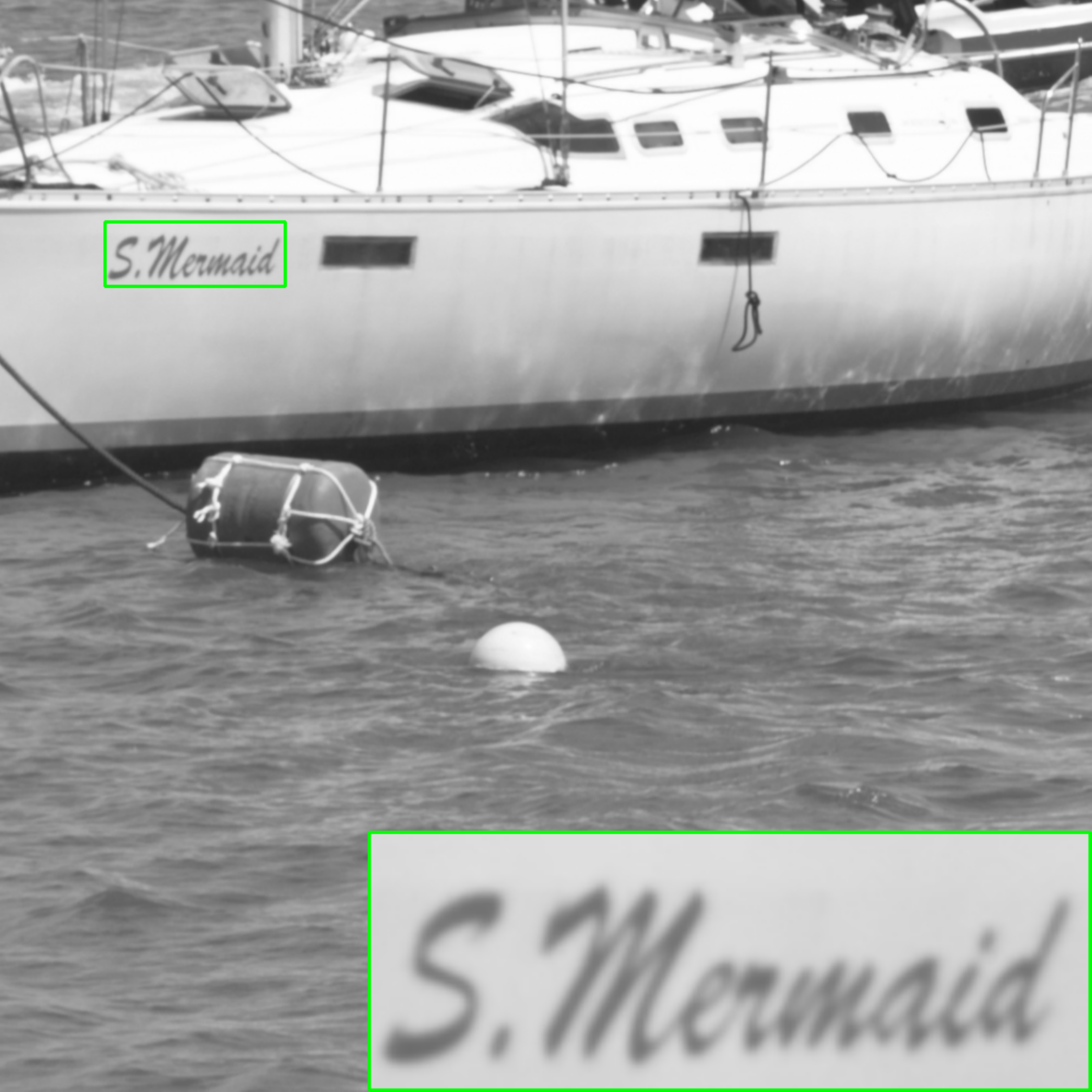}
    \caption{Ground Truth}
    \label{fig:intro-gt}
  \end{subfigure}  
  \caption{Reconstruction performance comparison on different methods. \cref{fig:intro-spike} presents one of the spike streams. \cref{fig:intro-gt} displays the ground truth image. \cref{fig:intro-tfi,fig:intro-tfp,fig:intro-spikeformer,fig:intro-Spk2ImgNet,fig:intro-WGSE,fig:intro-ours} are reconstruction results with different methods. From the enlarged images with characters, it becomes evident that our proposed method outperforms the competing methods, producing clear image with rich texture details for the high-speed object.} 
  \label{fig:intro}
\end{figure}

\section{Introduction}
\label{sec:intro}
The spike camera \cite{Alpher01}, which is an emerging neuromorphic camera, excels in high-speed motion capture with its high temporal resolution, low latency, low power consumption, and high dynamic range. These attributes are critical for high-speed imaging requirements. Aiming to solve the problems of motion blur and artifacts caused by the imaging mechanism that each pixel on the sensor of traditional digital camera is exposed synchronously in high-speed motion scenes, each pixel on the sensor of spike camera records the external absolute light intensity independently and continuously to generate spike streams with high temporal resolution. As a result, the spike camera can not only finely record the high-speed motion process, but also provide rich lighting information to reconstruct the texture details in the scene. 

However, converting binary spike streams to images requires specialized algorithms. Traditional methods \cite{Alpher02,Alpher05,Alpher06} perform basic reconstructions but struggle with noise and motion blur. Deep learning, notably CNNs and transformers, has advanced spike camera image reconstruction significantly. For example, Spk2ImgNet \cite{Alpher07} directly reconstructs images but may miss global context. SpikeFormer \cite{Alpher08}, leveraging transformers, excels but demands high computational resources for large images. WGSE \cite{Alpher09}, using discrete wavelet transforms, struggles with nonlinear data patterns. As depicted in 
 \cref{fig:intro}, none fully restores image details, highlighting the need to exploit spatial-temporal features in spike streams effectively.

In this paper, we propose Swin Spikeformer(SwinSF), a spike camera image reconstruction model designed to reconstruct dynamic scenes clearly from spike streams. In detail, SwinSF is made up of three main modules: Spike Feature Extraction Module, Spatial-Temporal Feature Extraction Module and Final Reconstruction Module. The Spike Feature Extraction Module uses convolution layers to extract spike features directly from spike streams. The Spatial-Temporal Feature Extraction Module consists of several Residual Swin Spikeformer Blocks (RSSB), which include a series of Spike Attention Blocks (SAB). By combining shifted window self-attention and proposed Temporal Spike Attention (TSA), SAB can fully extract both spatial and temporal information from both intra-frames and inter-frames, leading to a more robust and accurate reconstruction of spike streams. Ultimately, the Final Reconstruction Module utilizes a fusion of both spike and spatial-temporal features to reconstruct images.

Furthermore, effective network training demands a robust dataset of spike streams paired with ground truth images. Acquiring matched dynamic scene images from traditional and spike cameras is tough. PKU-Spike-HighSpeed \cite{Alpher02} lacks ground truths, while spike-REDS \cite{Alpher07}, though synthetic, doesn't capture high-speed motion well and its resolution ($250\times400$) undershoots modern spike camera standards ($1000\times1000$). We've thus developed an advanced simulator that precisely emulates high-resolution spike generation, synthesizing spike streams with its ground-truth images.

The main contributions of this paper are summarized as follows: 
\begin{itemize}
    \item We propose a spike camera image reconstruction model named Swin Spikeformer to reconstruct the dynamic scenes from spike streams clearly.
    \item We design a spike attention mechanism to effectively learn long-range spatial-temporal representation for spike streams. To the best of our knowledge, this is the first attempt to design a spike-used-only attention mechanism.
    \item We develop a spike camera simulator to generate synthesized spike streams and corresponding ground-truth images, matching the resolution of the latest spike camera.
    \item Experiments on both real-world and synthesized datasets demonstrate that the proposed method achieves state-of-the-art performance.
\end{itemize}

\section{Related Works}
\subsection{Image Reconstruction of Spike Camera}
Spike cameras' advancement has spurred various applications, with image reconstruction being pivotal. Recent methods, like TFI \cite{Alpher02}, TFP \cite{Alpher02}, Spk2ImgNet \cite{Alpher07}, SpikeFormer \cite{Alpher08}, and WGSE \cite{Alpher09}, have emerged. TFI leverages inter-spike intervals (ISIs) for reconstruction, foundational yet limited by quantization and dark current issues. TFP uses a moving time window to reduce noise but risks blur in fast scenes. Spk2ImgNet and SpikeFormer apply CNNs and transformers, respectively, for direct reconstruction, improving visuals. WGSE innovates with time-frequency analysis through discrete wavelet transforms. Despite progress, spatial-temporal representation learning remains an area for enhancement.

\subsection{Vision Transformers}
Recently, the Transformer architecture \cite{Alpher12} has gained significant interest in computer vision due to its NLP successes. Many transformer-based methods \cite{Alpher11,Alpher13,Alpher14,Alpher15,Alpher16,Alpher18,Alpher27} have been proposed for tasks like image classification \cite{Alpher11,Alpher16,Alpher22,Alpher23}, segmentation \cite{Alpher18,Alpher24,Alpher25,Alpher26}, and object detection \cite{Alpher13,Alpher28,Alpher29}. While effective at capturing long-range dependencies \cite{Alpher27,Alpher31}, studies \cite{Alpher32,Alpher33,Alpher34} suggest combining transformers with convolutions for better results. Transformers have also been applied to low-level tasks like image and video super-resolution \cite{Alpher35,Alpher38,Alpher39,Alpher42}, deblurring \cite{Alpher44}, and denoising \cite{Alpher43}. However, these methods are not directly applicable to spike camera data, and their use in this field remains underexplored. This study aims to propose a transformer-based method for reconstructing high-quality images from continuous spike streams.
\begin{figure}[tb]
  \centering
  \includegraphics[height=3.5cm]{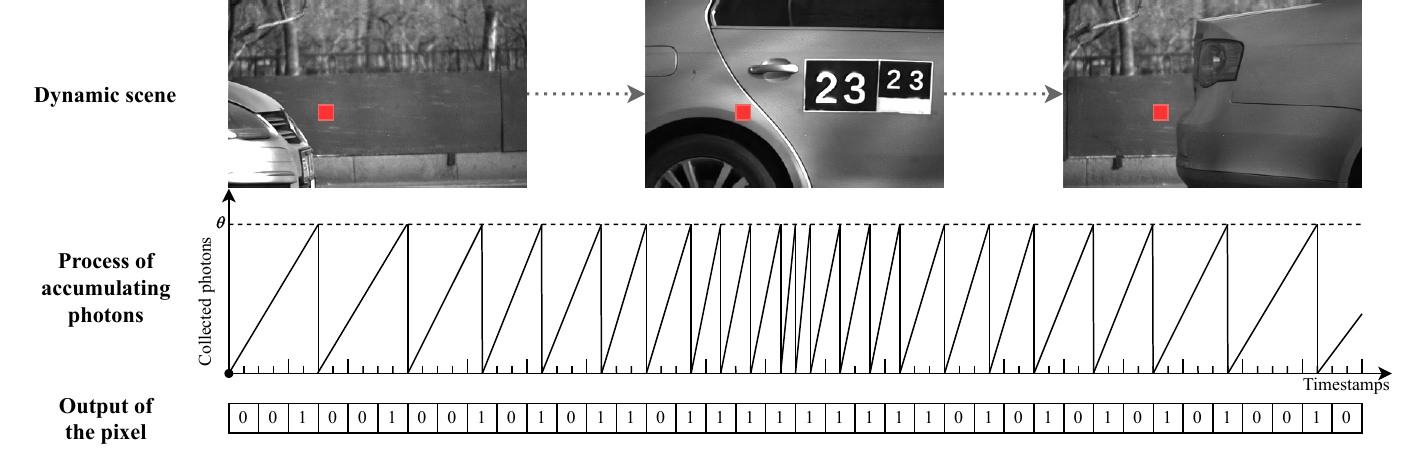}
  \caption{Illustration of the mechanism of how the spike camera accumulates photons and fires spikes when the light intensity changes due to the moving car.
  }
  \label{fig:mechanism}
\end{figure}
\section{Preliminaries: Mechanisms of Spike Camera}
The core of spike camera sensor is an asynchronous, persistently active pixel array. Each pixel independently accumulates photons; upon reaching a set threshold, it triggers a spike signal, resetting the count. This mechanism cycles continuously. Mathematically, this process is expressed as:
\begin{align}
A(i,j,t) = \int_{t_{(i,j)}^{pre}}^{t} \alpha I(i,j,t) \, dt \mod \theta,
\end{align}
where $A(i,j,t)$ denotes photon count at pixel $(i,j)$ at time $t$, $\alpha$ is the photoelectric rate, $I(i,j,t)$ is incident photon intensity. $\theta$ is the threshold.

Theoretically, pixels fire spikes anytime thresholds are met, but physically, the spike camera samples photon counts at high frequency, outputting discrete-time signals $S(i,j,t)$ at intervals $t = nT$, where $T$ is a short interval of microseconds. If a pixel $(i,j)$ meets the threshold at time $t = nT$, it reads out $S(i,j,t) =1$ and resets the count for the next cyclicality. Otherwise, it reads out $S(i,j,t) =0$. The above can be expressed as : 
\begin{align}
S(i,j,t) = 
\begin{cases} 
1 & \text{if } A(i,j,t) \geq \theta \\
0 & \text{if } A(i,j,t) < \theta
\end{cases}
.
\end{align}

Under continuous light, pixels independently accumulate photons, with high-speed sampling validating each pixel's spike status, forming an $H \times W$ spike frame. The sensor finally generates a sequence of spike frames called spike streams as the output of spike camera.

\cref{fig:mechanism} provides an illustrative example of a car moving through a scene. The background is darker than the car’s surface. Intuitively, the frequency of the spike is directly proportional to the brightness level observed, which is clearly evidenced by the red box depicted in the figure. 

\section{Method}

\subsection{Overall}

\begin{figure}[tb]
  \centering
  \includegraphics[height=7.5cm]{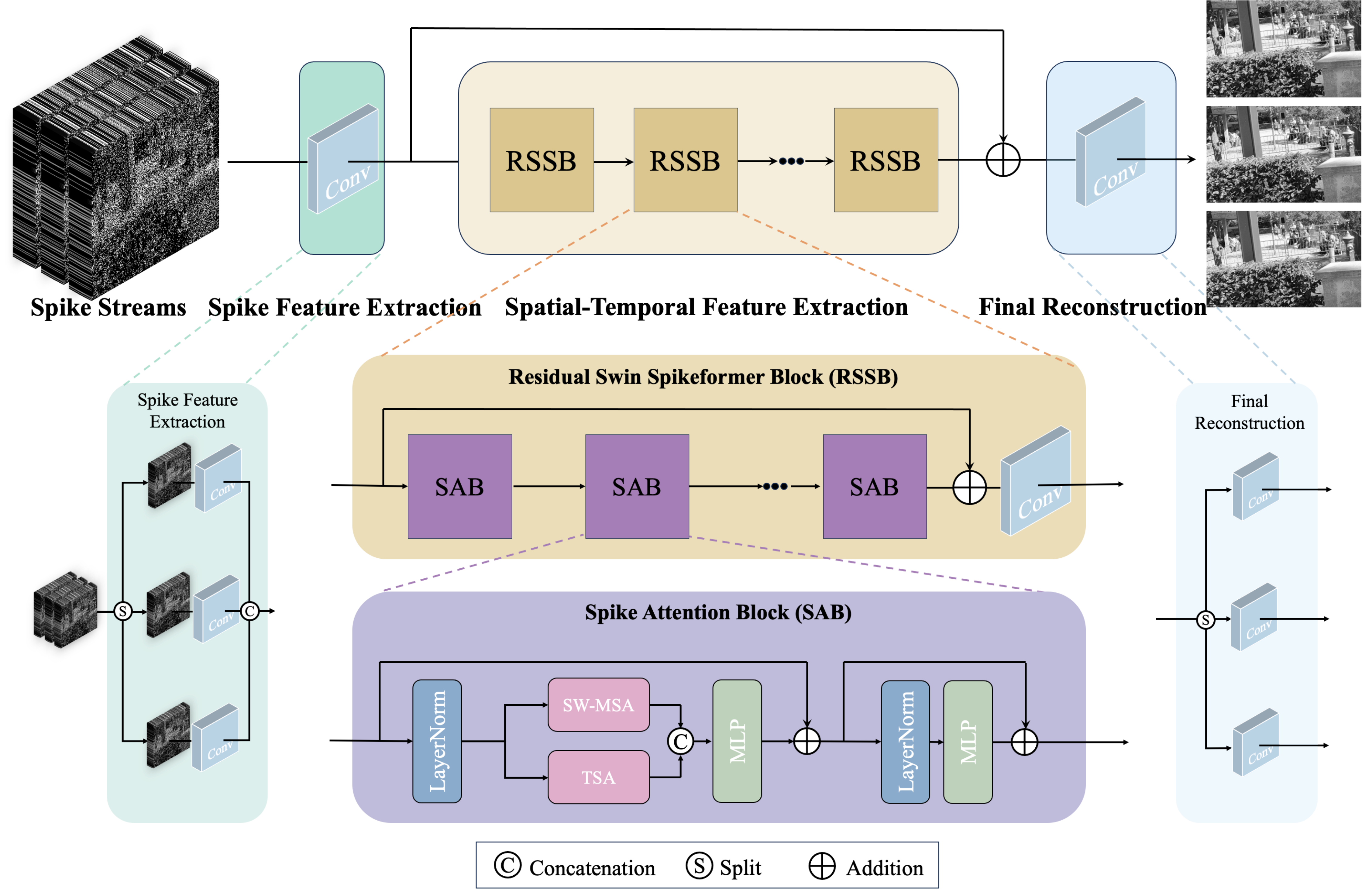}
  \caption{The framework of the proposed Swin Spikeformer(SwinSF), which is composed of Spike Feature Extraction, Spatial-Temporal Feature Extraction via Residual Swin Spikeformer Blocks(RSSB) with Temporal Spike Attention(TSA), and Final Reconstruction Module. Given the spike streams consist of a series of binary spike frames, SwinSF reconstructs the dynamic scenes for both the middle frame and its adjacent frames, which can fully exploit the temporal information to generate a high-quality reconstruction.}
  \label{fig:Overall}
\end{figure}
To fully learn spatial-temporal representation and reconstruct the dynamic scene from the spike streams, we propose a spike camera image reconstruction model called Swin Spikeformer(SwinSF). As shown in \cref{fig:Overall}, our network architecture is composed of three integral components: a Spike Feature Extraction Module, a Spatial-Temporal Feature Extraction Module, and a Final Reconstruction Module. This design draws upon the foundational elements established in prior research \cite{Alpher11}, yet innovates through its unique integration of these components. 

Specifically, given a binary consecutive spike stream $S \in \mathbb{B}_{\text{binary}}^{T \times H \times W}$ as input, we first split the spike stream into three parts in chronological order representing the left frame, middle frame, and right frame, respectively. Then, we extract the spike features $F_{sl}, F_{sm}, F_{sr} \in \mathbb{R}^{C \times H \times W}$ of each part by exploiting convolution layers, where $T$ denotes the temporal depth of the spike stream, $C$ denotes the channel number of the spike feature, and $H$ and $W$ denote the spatial dimensions. These extracted features are subsequently concatenated along the channel dimension, yielding $F_{s}\in \mathbb{R}^{3 \times C \times H \times W}$. Subsequently, a series of Residual Swin Spikeformer Blocks(RSSB) are utilized to perform the Spatial-Temporal Feature Extraction. Afterward, we add a global residual connection to fuse the concatenated spike features $F_{s}$ with spatial-temporal features $F_{ST}\in \mathbb{R}^{3 \times C \times H \times W}$. Finally, we reconstruct high-quality images via the Final Reconstruction module consisting of convolution layers, with each image representing the left frame, middle frame, and right frame respectively. As described in \cref{fig:Overall}, each RSSB contains several Spike Attention Blocks(SAB) and a convolution layer with a residual connection.
\subsection{Spike Attention Block(SAB)}
\label{sec:SAB}
As depicted in \cref{fig:mechanism}, the special imaging mechanism of the spike camera leads to a special dependence on the long-distance spatial-temporal information in the process of image reconstruction. Therefore, we design the Temporal Spike Attention(TSA) mechanism for spike streams to enhance the long-distance spatial-temporal information extraction ability of the network. The TSA block is inserted into the standard Swin Transformer Block \cite{Alpher11}, following the initial LayerNorm (LN) layer and in parallel with the shifted window-based multi-head self-attention (SW-MSA) module, as illustrated in \cref{fig:Overall}. The SW-MSA is employed periodically within consecutive TSA blocks, following a similar approach as described in \cite{Alpher11,Alpher39,Alpher42}. To reconcile the optimization and visual representation discrepancies between TSA and SW-MSA, we introduce a small constant $\beta$ as a scaling factor to the output of TSA. This adjustment harmonizes the interplay of the attention mechanisms, ensuring a more cohesive learning dynamic. The whole process of SAB is computed as:
\begin{equation}
\begin{aligned}
&\left\{
\begin{aligned}
&X_l, X_m, X_r = \text{LN}(\text{Split}(X)) \\
&Y_m = \text{SW-MSA}(X_m) + \beta \text{TSA}([X_l, X_m, X_r]) + X_m \\
&Y_m = \text{MLP}(\text{LN}(Y_m) + Y_m) \\
&Y = \text{Concat}(X_l, Y_m, X_r)
\end{aligned}
\right.,
\end{aligned}
\end{equation}
where $X$ denote the input feature, $X_l$, $X_m$, $X_r$, are three features split by $X$, representing the features at frames $l$,$m$, and $r$ respectively. $Y_{ms}$ and $Y_{mt}$ represent the intermediate features computed by the SW-MSA and TSA, and $Y$ is the output of the SAB.
\subsection{Temporal Spike Attention (TSA)}
To address the inherent limitation of SW-MSA in capturing solely the spatial visual representations within an isolated frame, and acknowledging the necessity for spatial-temporal feature extraction in image reconstruction of spike camera tasks, we introduce a novel mechanism called Temporal Spike Attention (TSA). This design is specifically tailored to distill spatial-temporal features across multiple frames, thereby harnessing the rich spatial-temporal information embedded within spike streams.

Given the middle frame features and the left and right adjacent frame features of size $C \times H \times W$, they are first partitioned into local windows of size $M \times M$, then TSA is calculated among the middle frame window features $X_m\in \mathbb{R}^{M^2 \times C}$ and the left and right adjacent frame window features $X_l, X_m \in \mathbb{R}^{M^2 \times C}$, $M^2$ is the number of window feature elements and $C$ is the channel number of feature. We compute the Query $Q$, Key $K$, Value $V$ from $X_l$, $X_m$, $X_r$ by linear projection as: 
\begin{align}
Q=X_l P_l, K=X_r P_r, V=X_m P_m,
\end{align}
where $P_l,P_m,P_r\in \mathbb{R}^{M^2 \times D}$ are projection matrices. $D$ is the channel number of projected features. Then we use $Q$ to query $K$ in order to generate the attention map $A = \text{SoftMax}\left(QK^T / \sqrt{D}\right) \in \mathbb{R}^{M^2\times M^2}$, which is then used for weighted sum of $V$, This is formulated as:
\begin{align}
TSA(Q,K,V) = \text{SoftMax}\left(QK^T / \sqrt{D}\right) V,
\end{align}
where \text{SoftMax($\cdot$)} means the row softmax operation.

\begin{figure}[tb]
  \centering
  \includegraphics[height=4.2cm]{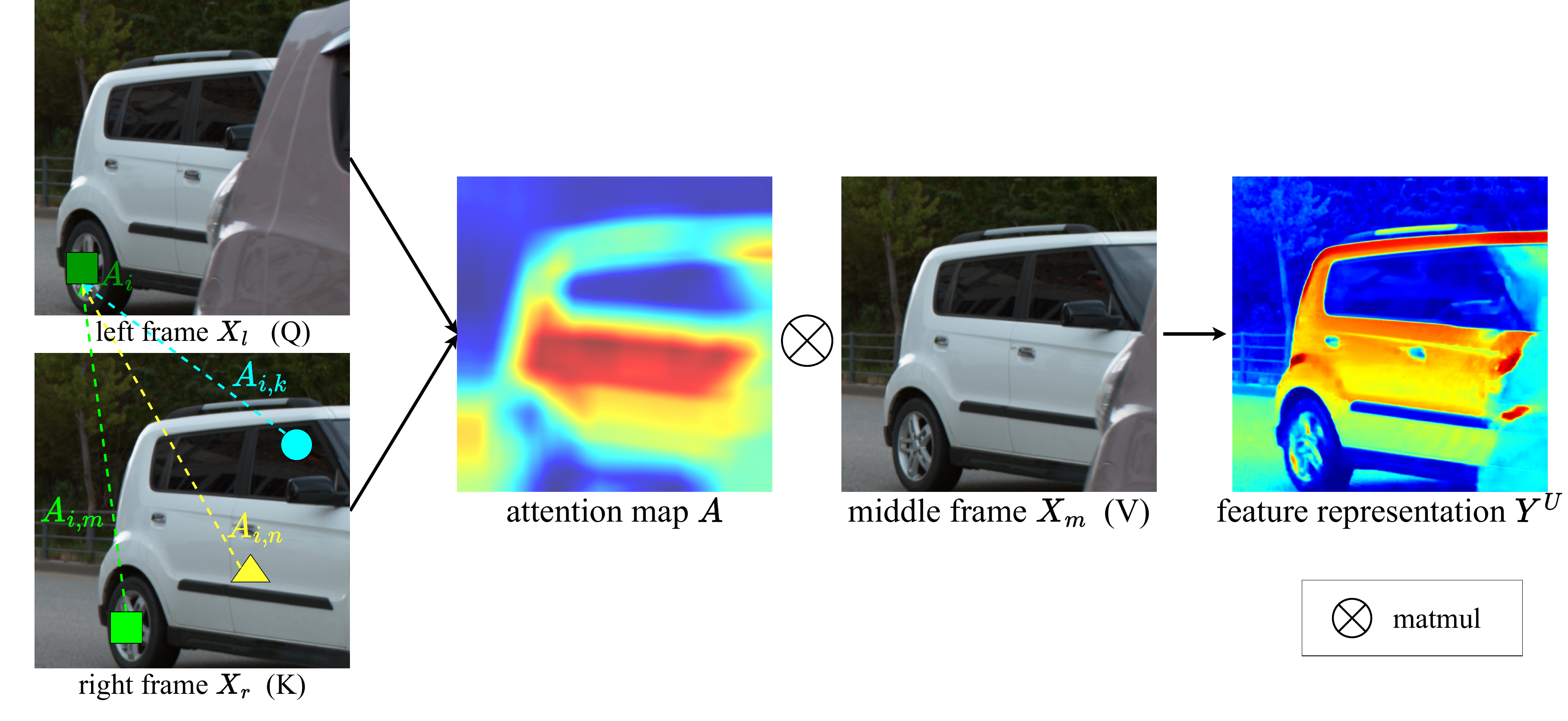}
  \caption{Illustrations for Temporal Spike Attention(TSA). We allow the dark green square(representing an element of the left frame) to query light blue circle, yellow triangle, and green square (representing elements of the right frame), utilizing their weighted features to form a new representation for the dark green square. The weights, depicted around dashed arrows (with only three examples shown for clarity), determine the influence of each queried element.
  }
  \label{fig:TSA}
\end{figure}

From the above process, we know that the TSA mechanism is adept at capturing the spatial-temporal dynamics inherent in spike streams. As illustrated in \cref{fig:TSA}, by computing the query and key from the adjacent frames, TSA is able to assess the spatial-temporal correlation between these frames, which is critical for the accurate reconstruction of the middle frame.
More specifically, the attention map $A$, generated from the interaction between $Q$ and $K$, reflects the degree of temporal similarity between the left and right frames. The updated middle frame feature representation $Y^U$ is computed as:
\begin{align}
Y^U = \sum_{i=1}^{N} A_i V_i,
\end{align}
where $A_i$ is the attention weight and $V_i$ is the value matrix. This formulation allows TSA to perform a weighted aggregation of features, where the weights are determined by the temporal correlation between the adjacent frames. By focusing on the temporal consistency between frames, TSA can effectively reconstruct the middle frame even in the presence of significant temporal sparsity. This is a distinct advantage over traditional attention mechanisms, which do not account for the temporal aspect of the spike streams and may struggle with the asynchronous nature of spike streams. The ability of TSA to leverage temporal information allows it to surpass these traditional methods, providing a more accurate and temporally coherent reconstruction of the visual scene.

\subsection{Loss Function}
Given that our network simultaneously reconstructs the dynamic scene of the middle spike frame and its adjacent spike frames, while effectively utilizing the inter-frame information in the reconstruction process, we employ a two-part loss function to optimize our network, which is formulated as:
\begin{equation}
\begin{aligned}
&\left\{
\begin{aligned}
&\text{Loss} = \lambda L_{\text{adj}} + L_{\text{mid}}  \\
&L_{\text{adj}} = \left\| I_l - G_l \right\|_1 + \left\| I_r - G_r \right\|_1 \\
&L_{\text{mid}} = \left\| I_m - G_m \right\|_1
\end{aligned}
\right.,
\end{aligned}
\end{equation}
where $I_i$ is the reconstructed dynamic scene of frame $i$, $G_i$ is the ground truth image of frame $i$ and $\lambda$ is the parameter balancing these two losses.
\subsection{Construction of New Dataset}
\label{tab:new_data}
To train our models, we need a dataset of spike stream inputs with corresponding ground truth images. The spike-REDS dataset \cite{Alpher07}, derived from the REDS dataset \cite{Alpher45}, is limited by its resolution ($250\times400$) and lack of high-speed motion representation. To address this, we developed an advanced spike camera simulator for $1000\times1000$ resolution. Using the X4K1000fps video dataset \cite{Alpher46} for high-resolution, fast-motion ground truth, we extract one or two $1000\times1000$ regions per frame. We apply the Super-SloMo frame interpolation algorithm \cite{Alpher47} to increase temporal resolution, simulating the high temporal acuity of a spike camera. Our simulator then generates high-fidelity spike streams. We created the spike-X4K dataset, comprising 1200 spike stream-ground truth pairs for training and 45 pairs for testing.

\section{Experiments}
\subsection{Datasets}
To evaluate our network, we conduct experiments on synthesized and real-life datasets with varying resolutions. For $250 \times 400$ resolution, models are trained on spike-REDS \cite{Alpher07} and tested on spike-REDS and PKU-Spike-HighSpeed \cite{Alpher02}. For $1000 \times 1000$ resolution, models are trained and tested on spike-X4K, as shown in \cref{tab:new_data}.
\subsection{Details}
We use 2 RSSB blocks, each with 6 SAB blocks. The temporal window sizes for adjacent and middle frames are 28 and 41, respectively. For training at $250 \times 400$ resolution, the spatial local window size is $5 \times 5$, head size 2, patch size 1, and channel size 96. At $1000 \times 1000$ resolution, the window size is $5 \times 5$, head size 1, patch size 4, and channel size 64.

Training is done with PyTorch using the Adam optimizer, starting with a learning rate of 0.0001. Models are trained for 900 epochs, with the learning rate halved every 300 epochs. Training is performed on 4 NVIDIA RTX3090 GPUs (24GB) with a batch size of 4 for $250 \times 400$ resolution, and on 1 NVIDIA Tesla V100 GPU (32GB) with a batch size of 1 for $1000 \times 1000$ resolution.
\begin{figure}[tb]
  \centering
  \includegraphics[height=4.2cm]{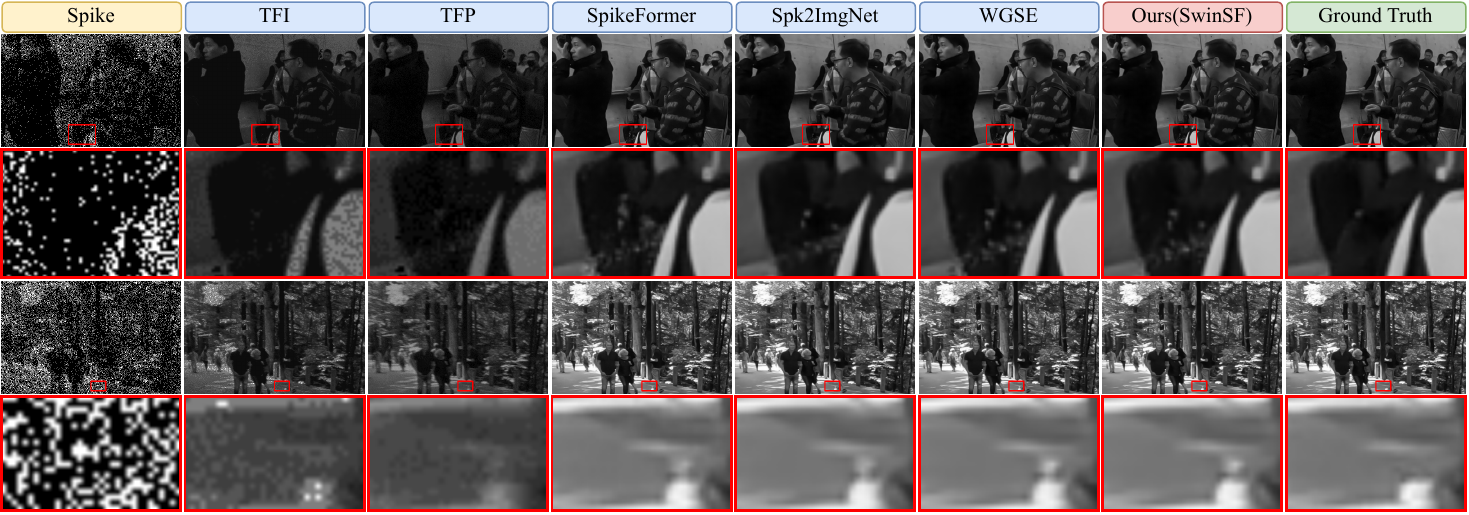}
  \caption{Reconstruction results on spike-REDS Dataset compared with other methods.
  }
  \label{fig:compare_reds}
\end{figure}
\subsection{Comparison With State-of-the-Art}
To evaluate our proposed network SwinSF, we compare it in different datasets and resolutions with recent works, i.e., TFP, TFI, SpikeFormer, Spk2ImgNet, WGSE. To be specific, WGSE was the state-of-the-art method before this.
\begin{table}[tb]
    \centering
    \caption{Quantitative comparison on spike-REDS, \textuparrow indicates the higher is better.}
    \begin{tabular}{c|p{1.2cm}p{1.2cm}p{1.8cm}p{1.8cm}p{1.5cm}|c}
    \toprule
    Model &TFI &TFP &SpikeFormer &Spk2ImgNet &WGSE &\textbf{Ours(SwinSF)} \\
    \midrule
        PSNR\textuparrow  &  24.94& 22.37 &37.18 &38.44 & 38.88 & \textbf{39.34} \\
        SSIM\textuparrow  &  0.7150 &  0.5801 &0.9738 & 0.9767&0.9774&  \textbf{0.9803} \\
    \bottomrule

    \end{tabular}
    
    \label{tab:res_on_reds}
\end{table}

\textbf{Quantitative and Qualitative Comparison on the resolution of 250 $\times$ 400}. 
When evaluating our network at a resolution of $250\times 400$, we train with spike-REDS and test on both spike-REDS and the PKU-Spike-HighSpeed Dataset. To quantitatively compare different reconstruction methods, we employ two full-reference image quality assessment (IQA) metrics, namely PSNR and SSIM, for synthesized data evaluation, and a no-reference IQA metric, NIQE, for real-world data assessment. As depicted in \cref{tab:res_on_reds}, our SwinSF outperforms other methods on the spike-REDS dataset, achieving the highest PSNR at 39.34dB with a notable gain of over 0.46dB, underscoring its superior effectiveness. Our SwinSF also achieves the highest SSIM at 0.9803. Furthermore, \cref{tab:res_on_real} reveals that SwinSF consistently delivers the best NIQE scores on PKU-Spike-HighSpeed dataset.
\begin{table}[tb]
 \caption{Quantitative comparison on PKU-Spike-HighSpeed, $\downarrow$ indicates the lower is better.}
    \centering
    \begin{tabular}{c|p{1.2cm}p{1.2cm}p{1.8cm}p{1.8cm}p{1.5cm}|c}
    \toprule
    Model &  TFI& TFP&SpikeFormer&Spk2ImgNet &WGSE &\textbf{Ours(SwinSF)} \\
    \midrule
        NIQE$\downarrow$  &  9.1843& 8.1286 &5.6734 &4.8864 & 5.3276 & \textbf{4.5814} \\
    \bottomrule

    \end{tabular}
   
    \label{tab:res_on_real}
\end{table}
\begin{figure}[tb]
  \centering
  \includegraphics[height=5.3cm]{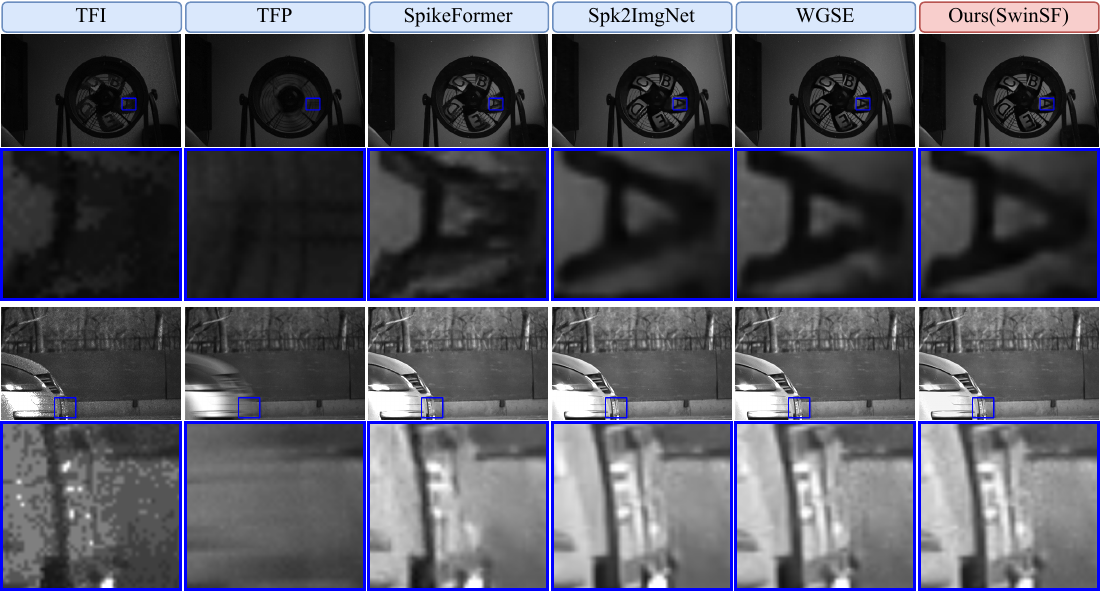}
  \caption{Reconstruction results on PKU-Spike-HighSpeed compared with other methods.
  }
  \label{fig:compare_classA}
\end{figure}

\cref{fig:compare_reds,fig:compare_classA} present the reconstruction results from various methods for real-world and synthesized data. The reconstructions by our SwinSF are visually superior to those of competing approaches.
Our SwinSF consistently delivers high-quality reconstructions, effectively revealing crisp textures and intricate details even within high-speed motion scenes. 
Due to the relatively simplistic construction and outdated resolution of spike-REDS, we place particular emphasis on the evaluation of spike-X4K.
\begin{table}[tb]
    \centering
    \caption{Quantitative comparison on spike-X4K, \textuparrow indicates the higher is better.}
    \begin{tabular}{c|p{1.2cm}p{1.2cm}p{1.8cm}p{1.8cm}p{1.5cm}|c}
    \toprule
    Model &  TFI& TFP&SpikeFormer&Spk2ImgNet &WGSE &\textbf{Ours(SwinSF)} \\
    \midrule
        PSNR\textuparrow  &  15.84& 15.87 &36.84 &37.95 & 38.19 & \textbf{39.61} \\
        SSIM\textuparrow  &  0.6420 &  0.7064 &0.9474 & 0.9519&0.9524&  \textbf{0.9682} \\
    \bottomrule
    \end{tabular}
    \label{tab:res_on_x4k}
\end{table}

\begin{figure}[tb]
  \centering
  \includegraphics[height=7.5cm]{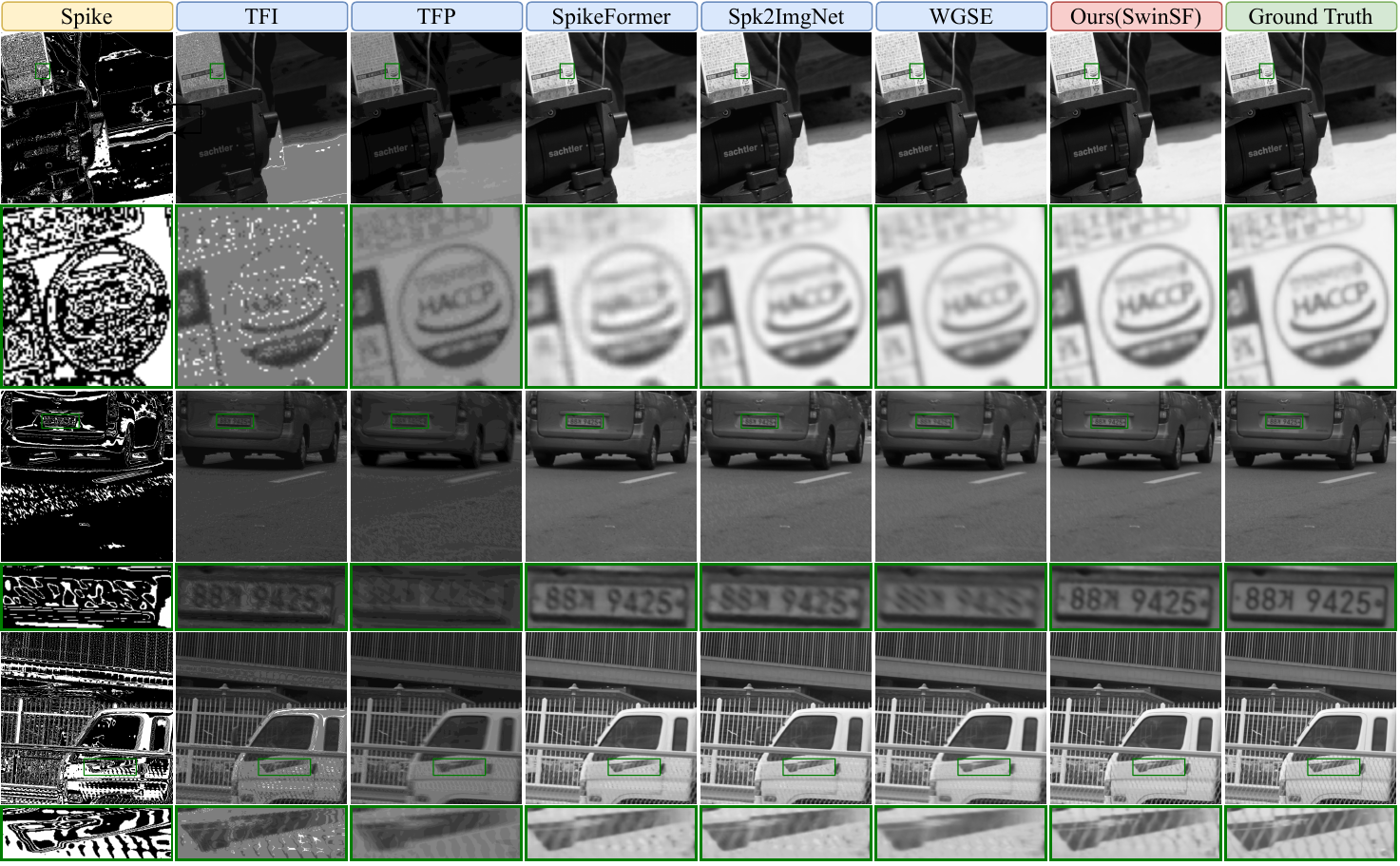}
  \caption{Reconstruction results on spike-X4K Dataset compared with other methods.
  }
  \label{fig:compare_x4k}
\end{figure}

\textbf{Quantitative and Qualitative Comparison on the resolution of 1000$\times$ 1000}.
When evaluating our network at the resolution of $1000\times 1000$, we train with spike-X4K and test on spike-X4K. To quantitatively compare different reconstruction methods, we also employ PSNR and SSIM for synthesized data evaluation. As depicted in \cref{tab:res_on_x4k}, our SwinSF outperforms other methods on the spike-X4K dataset, achieving the highest PSNR at 39.61 and the highest SSIM at 0.9682. \cref{fig:compare_x4k} presents the reconstruction results from various methods for spike-X4K. While other methods may capture the contours of swiftly moving objects, their reconstructions often appear noisy and lack fine texture detail. In stark contrast, our SwinSF consistently delivers stable reconstruction quality, preserving clear textures and intricate details, even within scenes characterized by high-speed motion.
\section{Ablation Study}
\subsection{Effectiveness of SAB}
To evaluate SAB's impact, we conducted ablation studies on spike-REDS. $V_1$, a baseline, mirrors a standard transformer layer \cite{Alpher12} with only Multihead Self-Attention (MSA). $V_2$ adopts Shifted Window Multihead Self-Attention (SW-MSA) \cite{Alpher11}, akin to a basic Swin Transformer layer. $V_3$ focuses solely on Temporal Spike Attention (TSA), eschewing spatial attention. $V_4$ combines MSA and TSA for a dual spatial-temporal attention approach. $V_5$, our proposed configuration of SAB, further builds upon $V_2$ by adding TSA to the SW-MSA framework.
\begin{table}[tb]
    \centering
    \caption{Effectiveness of the proposed SAB, \textuparrow indicates the higher is better.}
    \begin{tabular}{c|>{\centering\arraybackslash}p{1.5cm}>{\centering\arraybackslash}p{1.5cm}>{\centering\arraybackslash}p{1.5cm}>{\centering\arraybackslash}p{1.5cm}>{\centering\arraybackslash}p{1.5cm}}
    \toprule
    Metric &  $V_1$& $V_2$&$V_3$&$V_4$&$V_5$\textbf{(Ours)}\\
    \midrule
    MSA&$\checkmark$ & & & $\checkmark$ & \\
    SW-MSA& & $\checkmark$ & & & $\checkmark$ \\
    TSA & & & $\checkmark$ &$\checkmark$ & $\checkmark$ \\
    \midrule
        PSNR\textuparrow  &  36.02 & 38.65 & 36.07 & 36.82 & 39.34 \\
        SSIM\textuparrow  &  0.9588 & 0.9754 & 0.9597 & 0.9683 & 0.9803 \\
    \bottomrule
    \end{tabular}
    \label{tab:abl_sab}
\end{table}
Analyzing \cref{tab:abl_sab}, $V_5$ outperforms $V_2$ by 0.69, while $V_4$ exceeds the performance of V1 by 0.80. This confirms Temporal Spike Attention (TSA)'s effectiveness. $V_3$'s 0.05 improvement over $V_1$ stresses temporal information's importance in spike data reconstruction. $V_2$'s superiority over $V_1$ and $V_5$ over $V_4$ underscores the Swin Transformer's \cite{Alpher11} positive effect. It highlights the Swin Transformer's capacity to boost performance and collaborate with TSA for spatial-temporal information mining.
\begin{table}[tb]
    \centering
    \caption{Effectiveness of the scaling factor $\beta$ of TSA, \textuparrow indicates the higher is better.}
    \begin{tabular}{c|>{\centering\arraybackslash}p{1.5cm}>{\centering\arraybackslash}p{1.5cm}>{\centering\arraybackslash}p{1.5cm}>{\centering\arraybackslash}p{1.5cm}}
    \toprule
    $\beta$ & 0 & 1 & 0.5 & 0.1\\
    \midrule
    PSNR\textuparrow  &38.65&39.21&39.28&39.34\\
    SSIM\textuparrow  &0.9754&0.9767&0.9784&0.9803\\
    \bottomrule
    \end{tabular}
    
    \label{tab:abl_tsa}
\end{table}
\begin{table}[tb]
    \centering
    \caption{Effectiveness of the parameter $\lambda$ of loss function, \textuparrow indicates the higher is better.}
    \begin{tabular}{c|>{\centering\arraybackslash}p{1.5cm}>{\centering\arraybackslash}p{1.5cm}>{\centering\arraybackslash}p{1.5cm}>{\centering\arraybackslash}p{1.5cm}}
    \toprule
    $\lambda$ & 0 & 1 & 0.5 & 0.1\\
    \midrule
    PSNR\textuparrow  &39.05&39.23&39.31&39.34\\
    SSIM\textuparrow  &0.9773&0.9786&0.9795&0.9803\\
    \bottomrule
    \end{tabular}
    \label{tab:abl_loss}
\end{table}
\subsection{Effectiveness of the scaling factor of TSA}
We conduct experiments to explore the effects of the scaling factor $\beta$ of TSA. As presented in the manuscript 
 \cref{sec:SAB}, $\beta$ is used to control the weight of TSA features for feature fusion. A larger $\beta$ means a larger weight of features extracted by TSA and $\beta = 0$ represents TSA is not used. As shown in \cref{tab:abl_tsa}, the model with $\beta = 0.1$ obtains the best performance. This observation suggests a nuanced interplay between TSA and SW-MSA, potentially hinting at optimization challenges when both are concurrently employed.  Crucially, a minimal $\beta$ weighting for the TSA component appears to mitigate these challenges, facilitating a more harmonious integration of temporal and spatial attention mechanisms.
\subsection{Effectiveness of the parameter of loss function}
To investigate the impact of varying the parameter $\lambda$ of the loss function, we established a range of $\lambda$ values to assess changes in performance. As illustrated in \cref{tab:abl_loss}, it is observed that the loss function achieves optimal performance when $\lambda$ is set to 0.1. In contrast, setting $\lambda$ to 0.5 or 1 results in no significant improvement in model performance, and may even lead to a decline.

\subsubsection{\ackname}
This work was supported by the National Key R\&D Program of China (2021ZD0109802) and supported by High-performance Computing Platform of BUPT.

%
%
\bibliographystyle{splncs04}
\bibliography{main}
\end{document}